\pdfoutput=1

\documentclass[11pt]{article}
\usepackage{EMNLP2023}

\usepackage{times}
\usepackage{array}
\usepackage{colortbl}
\usepackage{booktabs}
\usepackage{latexsym}
\usepackage{tcolorbox}
\usepackage{mathtools}
\usepackage[T1]{fontenc}

\usepackage[utf8]{inputenc}

\usepackage{microtype}

\usepackage{inconsolata}

\title{Chain of Stance: Stance Detection with Large Language Models}

\author{Junxia Ma$^1$, Changjiang Wang$^1$, Hanwen Xing$^1$, Dongming Zhao$^1$, Yazhou Zhang$^2{}^*$\\
  $^1$College of Software Engineering, Zhengzhou University of Light Industry \\
  $^2$Tianjin University \\
  \texttt{yzzhang@zzuli.edu.cn}
}

\begin{document}
\maketitle
\begin{abstract}
Stance detection is an active task in natural language processing (NLP) that aims to identify the author's stance towards a particular target within a text. Given the remarkable language understanding capabilities and encyclopedic prior knowledge of large language models (LLMs), how to explore the potential of LLMs in stance detection has received significant attention. Unlike existing LLM-based approaches that focus solely on fine-tuning with large-scale datasets, we propose a new prompting method, called \textit{Chain of Stance} (CoS). In particular, it positions LLMs as expert stance detectors by decomposing the stance detection process into a series of intermediate, stance-related assertions that culminate in the final judgment. This approach leads to significant improvements in classification performance. We conducted extensive experiments using four SOTA LLMs on the SemEval 2016 dataset, covering the zero-shot and few-shot learning setups. The results indicate that the proposed method achieves state-of-the-art results with an F1 score of 79.84 in the few-shot setting. 
\end{abstract}

\section{Introduction}
Recently, the rapid development of social media platforms such as X, WeChat, and TikTok has led to an explosive growth in user-generated content. The increasing availability of massive amounts of textual data has created a need to automatically analyze the opinions, emotions, and stances expressed. This is because understanding the stance tendencies of the public offers significant benefits for political analysis, public opinion polling, and rumor detection. Hence, stance detection has been an active topic in NLP. Stance detection aims to identify the author's attitude or stance towards a specific target (such as an entity, concept, or event)\cite{Mohammad}, typically classified into categories such as support, opposition, or neutrality. 

\begin{figure}
    \centering
    \includegraphics[width=1\linewidth]{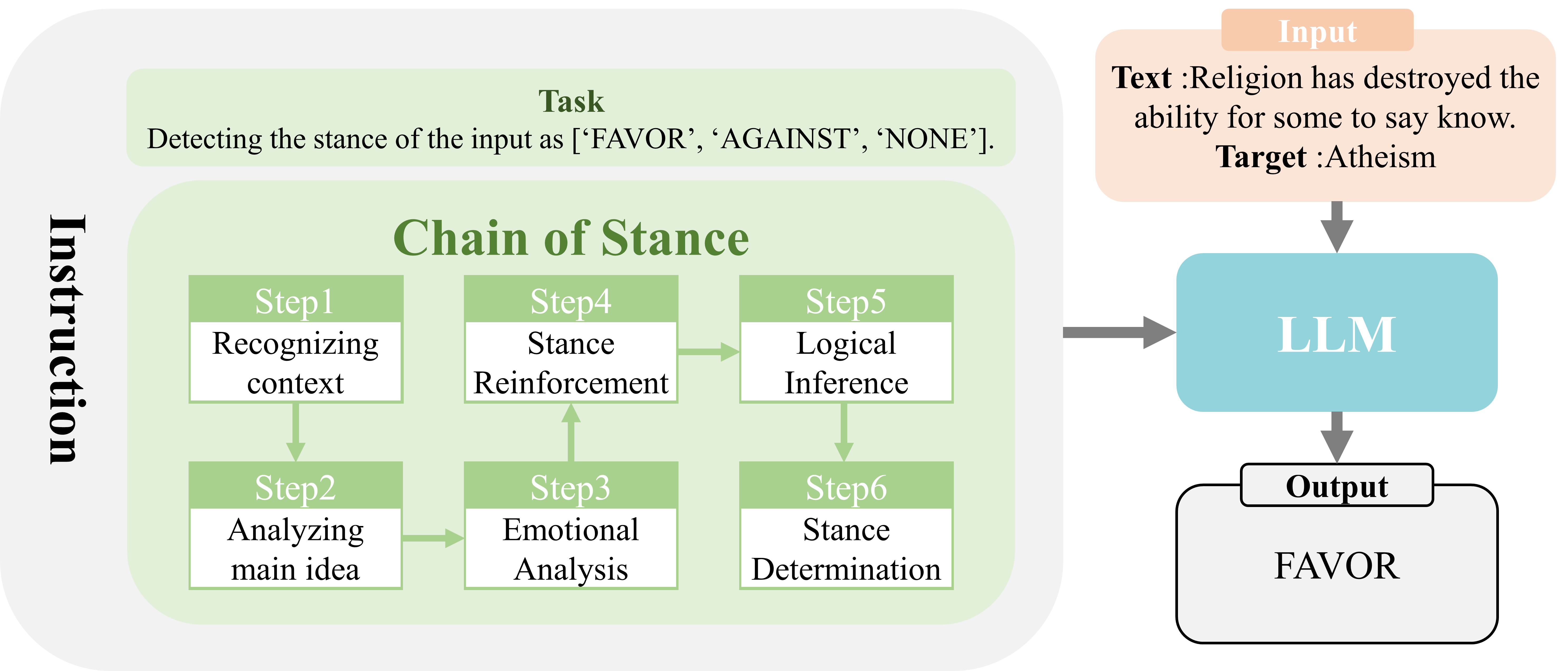}
    \caption{The overall architecture of CoS.}
    \label{img1}
\end{figure}

Recently, the advents of LLMs, e.g., ChatGPT\footnote{https://chat.openai.com/}, GPT-4\cite{gpt4}, Qwen2\cite{qwen}, LLaMA2~\cite{llama}, etc., have demonstrated their outstanding performance across  numerous NLP tasks. Through instruction fine-tuning and in-context learning, LLMs have possessed marvelous language understanding, generation and reasoning abilities. Now, their transformative momentum has surged into the task of stance detection, giving rise to a nascent wave of stance detection LLMs.  For instance, wang et al.\cite{deem} proposed a Dynamic Experienced Expert Modeling(DEEM) method to generate experienced expert. Li et al.\cite{15} designed a gated calibration network to mitigate the biases on the stance reasoning results from LLMs. However, the above-mentioned works only focused on fine-tuning the LLMs with complex samples, with high computing and time costs. 

To fill this gap, we propose a novel stance detection method called chain of stance (CoS). This approach leverages the encyclopedic prior knowledge of LLMs to detect human's stance. Inspired by CoT~\cite{4} that builds a sequence of intermediate reasoning steps to improve the reasoning ability of LLMs in math area, CoS also adopts this sequential structure. But the key difference is that we capture the relation between human's emotion and his/her stance, instead of letting LLMs think freely. In particularly, it positions LLMs as expert stance detectors by decomposing the stance detection process into a series of intermediate, stance-related assertions that culminate in the final judgment. This approach leads to significant improvements in classification performance. In this work, \textit{assertion} is similar to \textit{thought}, which is a coherent language sequence related to context, emotion, or opinion that serves as an intermediate indicator toward identifying stance.

Finally, we present empirical evaluations of the proposed CoS method across the SemEval 2016 dataset over 4 SOTA LLMs (i.e., Mistral-7B, Qwen 1.5-7B, LLaMA 3-8B, LLaMA 2-7B), and compare their results against 7 SOTA baselines.  
The proposed chain of stance (CoS) method achieved state-of-the-art results, with a leading F1 score of 76.43 in zero-shot and 79.84 in few-shot settings.  Further error analysis also presents its experimental results. 

The main contributions are concluded as follows:
\begin{itemize}
\item Our research is the first to employ chained reasoning approach for stance detection.
\item We propose a new prompting framework, chain of stance, which models the stance detection process as a sequence of stance-related assertions.
\item The experimental results prove the superiority and broad applicability of the proposed prompting framework.
\end{itemize}

\section{Related Work}
\subsection{Stance Detection}
Stance detection has long been an active research area in the field of Natural Language Processing (NLP). Early studies employed various machine learning and deep learning techniques to tackle this task\cite{5}.In recent years, the advent of large language models (LLMs) has marked a significant advancement in artificial intelligence, pushing the state-of-the-art performance on many stance detection datasets\cite{Li}. Models like GPT-3.5, which have demonstrated powerful capabilities in numerous areas of NLP, have garnered considerable attention. Trained on extensive datasets, these models have significantly improved their ability to understand and mimic human language patterns\cite{4}. Researchers are actively utilizing the potenial of LLMs for stance detection. For instance, Mets et al.\cite{mets2024automated} examined the applicability of LLMs in automated stance detection within challenging contexts. Wang et al.\cite{deem} leveraged the generation capability of LLMs to simulate specific experts (i.e.,multi-agents) to detect the stance. Ilker et al.\cite{stance_fine_tune} argue that large language models (LLMs) possess exceptional capabilities in stance detection and call for more extensive research in this area. 

\subsection{Prompt Tuning}
Prompt learning transforms downstream learning tasks into text generation tasks by incorporating prompt information into the text input~\cite{7}. Prompt tuning has proven to effectively enhance the capabilities of large language models (LLMs) in many NLP tasks, such as sentiment analysis and text classification, and is therefore widely adopted. Hardalov et al.~\cite{hardalov} designed a prompt-based framework for cross-language stance detection. Ding et al.~\cite{ding2024leveraging} utilizes a chain-of-thought approach to elicit knowledge and fuses knowledge through a multi-prompt learning network. Zhu et al.~\cite{zhu2024short} introduced soft knowledge during cue fine-tuning, a strategy that effectively enhances the model's understanding of the context of short texts and significantly improves text categorization. Hu et al.~\cite{huang2023knowledge} expands the verbalizer in prompt-tuning using external semantic knowledge and infusing background knowledge. Wei et al.~\cite{4} made a formal definition of CoT prompting in LLMs and proved its effectiveness by presenting empirical evaluations on arithmetic reasoning benchmarks. This work pioneered the use of CoT prompting in NLP.

However, most existing stance detection methods do not explicitly model the reasoning process, making the model's predictions difficult to interpret. Additionally, these methods typically treat stance detection as a binary or multi-class classification task\cite{deem}, ignoring the inherent complexity of the problem. 

\section{Methodology}
\subsection{Task Definition}
The task of stance detection is defined as follows: Given a text $S$ and a target $t$, the model needs to determine the stance polarity $y$ towards $t$, such as \textit{favor}, \textit{against}, and \textit{none}. We now use a chain of stance prompt-based approach to solve the stance detection task. Specifically, we use the following prompt template as the input to the LLM:

\begin{tcolorbox}[colback=gray!10!white, colframe=gray!50!black, width=\columnwidth, arc=0mm, boxrule=0mm]
      Given the text $S$ and the chain of stance $C$, what is the stance polarity towards $t$?
\end{tcolorbox}

Specifically, the LLM returns the answer using Formula 1. 
\begin{equation}
\hat{y}=\operatorname{argmax} p(y \mid S, C, t)
\end{equation}

\subsection{The Proposed Approach: CoS}
We propose a method called chain of stance(CoS), a CoT-style paradigm that allows LLMs to decompose the problem of stance detection into intermediate steps and solve each before making decision (Fig.\ref{img1}). Each step will be sampled sequentially, then the output is summarized based on the above steps.

Unlike traditional stance detection input/output prompts, we do not require the LLM to directly provide the stance detection result. Instead, we ask the LLM to consider various aspects and fully account for potential information before giving the final result to enhance detection accuracy. Specifically, CoS involves the following steps: 

\textbf{Step 1.} We first use the following template to help the LLM understand the contextual information $I$ of the text. 
\begin{tcolorbox}[colback=gray!10!white, colframe=gray!50!black, width=\columnwidth, arc=0mm, boxrule=0mm]
Given text $S$, understand the contextual information $i$ of the text, which includes the topic of the text, the identity of the author, the target audience, and the relevant socio-cultural background.
\end{tcolorbox}
This step can be formulated as $I=\operatorname{argmax} p(i \mid S, t)$, where $I$  is the contextual information contained in the text $S$. 

\textbf{Step 2.} Next, based on $S$, $t$, and $i$, we ask the LLM to interpret the main idea $V$ of the text $S$.
\begin{tcolorbox}[colback=gray!10!white, colframe=gray!50!black, width=\columnwidth, arc=0mm, boxrule=0mm]
What are the core viewpoints and main intentions to be expressed in the text?
\end{tcolorbox}
This step can be formulated as $V=\operatorname{argmax} p(v \mid S, t, i)$, where $V$ are the main ideas in the text $S$.

\textbf{Step 3.} In this step, we instruct the LLM to analyze the language expression and emotional attitude $E$ of the text. 
\begin{tcolorbox}[colback=gray!10!white, colframe=gray!50!black, width=\columnwidth, arc=0mm, boxrule=0mm]
Analyze the language expression and emotional inclination of the text. Identify the emotional words and rhetorical devices used within the text, and analyze the tone adopted by the author (e.g., affirmative, negative, interrogative, exclamatory, etc.). Based on this analysis, describe the author's emotional attitude $E$. 
\end{tcolorbox}
This step can be formulated as $E=\operatorname{argmax} p(e \mid S, t, i, v)$, where $E$  is the emotional attitude  expressed in the text $S$.

\textbf{Step 4.} In this step, we let the LLM compare the text $S$ with three possible stances (e.g., favor, against, none), and obtain the similarities and contrasts between the text and each potential stance. For each stance, calculate its probability given $S$, $t$, $i$, $v$, $e$.
\begin{tcolorbox}[colback=gray!10!white, colframe=gray!50!black, width=\columnwidth, arc=0mm, boxrule=0mm]
Compare similarities and contrasts between text $S$ and various possible stances (favor, against, none), based on the above-mentioned information.  
\end{tcolorbox}
The calculated probabilities for each stance are combined to form a set $A$: $A=\{\mathrm{favor}{:} P(favor|S,t,i,v,e), \mathrm{against}{:} P(against|\\S,t,i,v,e), \mathrm{none}{:} P(none|S,t,i,v,e)\}$.

\textbf{Step 5.} We ask the LLM to confirm the consistency and rationality of the stance.
\begin{tcolorbox}[colback=gray!10!white, colframe=gray!50!black, width=\columnwidth, arc=0mm, boxrule=0mm]
Conduct logical inference based on the context and other relevant information to confirm the consistency and rationality of the stance.
\end{tcolorbox}

This step can be formulated as $L=\operatorname{argmax} p(l \mid S, t, i, v, e, a)$, where $L$ is the  result of logical reasoning performed by the LLM.

\textbf{Step 6.} With the complete stance skeleton ($S$, $t$, $i$, $v$, $e$,  $a$, $l$) as context, we ask the LLM to make the final decision.
\begin{tcolorbox}[colback=gray!10!white, colframe=gray!50!black, width=\columnwidth, arc=0mm, boxrule=0mm]
Based on the above information, determine the stance polarity towards $t$.
\end{tcolorbox}
We note this step as $\hat{y}=\operatorname{argmax} p(y \mid S, t, i, v, e, a, l)$.

\begin{figure*}
    \centering
    \includegraphics[width=1\linewidth]{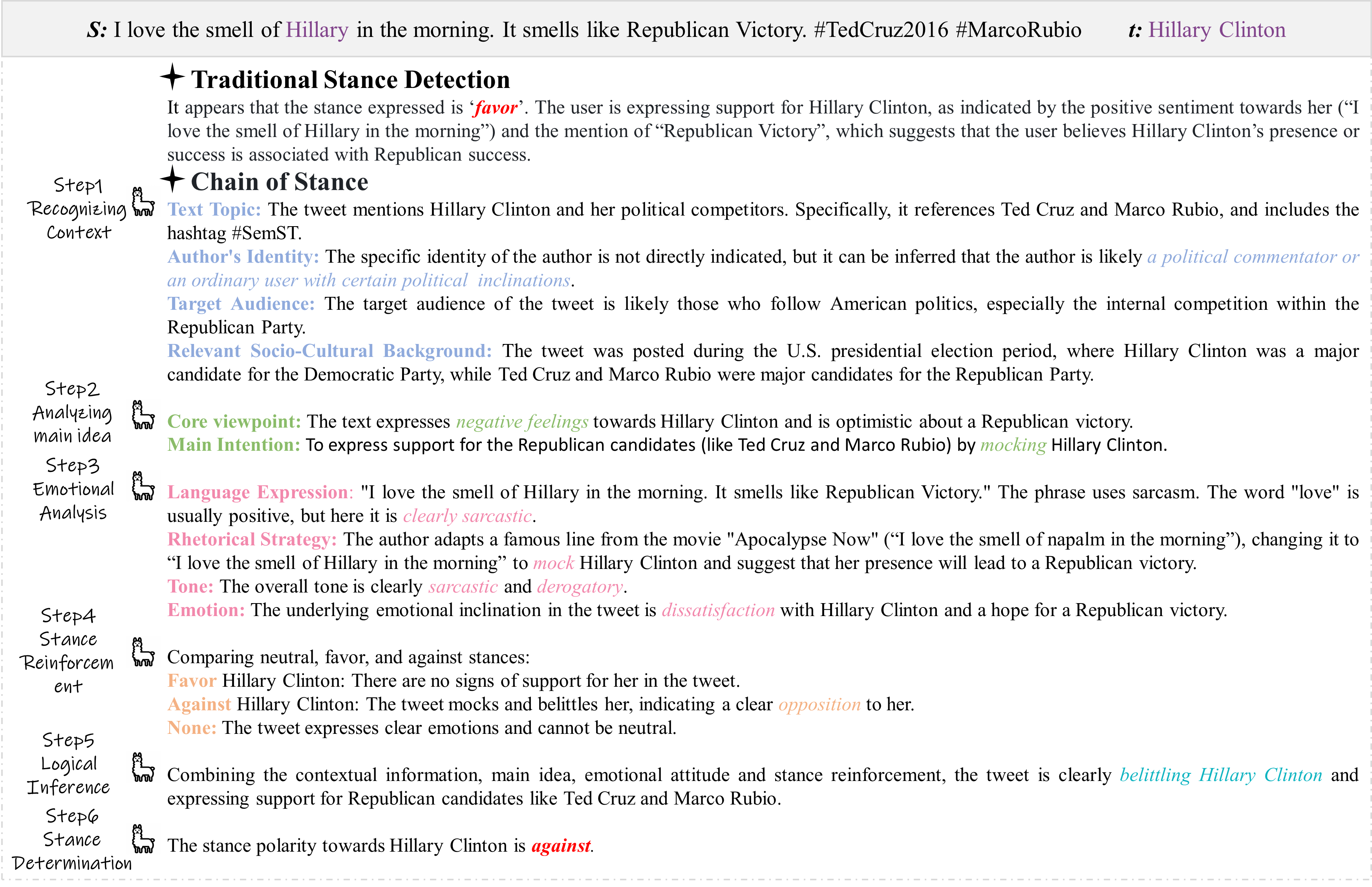}
    \caption{The specific implementation process of CoS.}
    \label{fig:enter-label}
\end{figure*}

By following these steps, the model acts like an experienced "stance expert", thoroughly extracting valuable information from the text while breaking down the stance detection process into a series of sequential steps. This guides the model to detect the stance of the text according to predefined procedures. This approach helps LLMs more comprehensively and meticulously understand and determine the author's stance. Such analysis enables a deeper understanding of the author's intentions and viewpoints, providing a foundation for further analysis and discussion. This not only enhances the accuracy of LLMs in stance detection but also improves the interpretability and rationale of the model's stance detection process.

\section{Experiment Settings}
\subsection{Dataset and Evaluation Metrics}
\textbf{Dataset}. SemEval-2016 Task 6~\cite{Mohammad}, named ``Stance Detection in Tweets'', is a benchmarking dataset widely used to explore and analyze users' stances on certain targets based solely on the content of their tweets. This dataset contains a total of 4,163 tweets, featuring five targets from different domains: Hillary Clinton (HC), Feminist Movement (FM), Legalization of Abortion (LA), Atheism (A), and Climate Change is a Real Concern (CC). The dataset includes three different stance labels: \textit{favor}, \textit{against}, and \textit{none}.

We use the evaluation metric recommended by the dataset creators for a fair comparison, where $F_{a v g}$ is the primary evaluation metric. $F_{a v g}$ is the average of the F1 scores for the ``favor'' and ``against'' labels. The specific calculation method for the evaluation metric is as follows: 

\begin{equation}
F_{\text {avg }}=\frac{F_{\text {favor }}+F_{\text {against }}}{2}
\end{equation}

\begin{table}
\caption{Statistics of SemEval-2016 Task6 dataset}
\label{table:1}
    \centering
    \footnotesize
    \setlength{\tabcolsep}{5pt} 
    \begin{tabular}{@{}lccccc@{}}
         \toprule 
         \textbf{Target}& \textbf{Train}& \textbf{Test}& \textbf{AGAINST}& \textbf{FAVOR}& \textbf{NONE}\\
         \midrule 
         HC&  689&  295&  565&  163& 256\\
         FM&  664&  285&  511&  268& 170\\
         LA&  653&  280&  544&  167& 222\\
         A&  513&  220&  464&  124& 145\\
         CC&  395&  169&  335&  26& 203\\
         \bottomrule 
    \end{tabular}
\end{table}

\subsection{Implementation Details}
We conducted experiments using four models: LLaMa2-7b-Chat, LLaMa3-8b-Instruct, Mistral-7b-Instruct-v0.2, and Qwen1.5-7b-Chat. For each model, we used the same instructions. 
To detail the fine-tuning parameters, the selected hyperparameters for LoRA included a rank of 8, an alpha of 16, and a dropout rate of 0.1. We trained each model with a batch size of 2 for a maximum of 10 epochs. The learning rate was set to 5e-5, and the compute type was fp16. We report averaged scores of 3 runs to obtain statistically stable results. The training was conducted using a single NVIDIA V100 GPU with 32GB of RAM. 

\subsection{Baseline Models}
To fully demonstrate the effectiveness of CoS, we selected 7 SOTA stance detection models as baselines for our experiments. They are:

\textbf{(1) JointCL \cite{11}\textit{.}} This model designs a new joint contrastive learning framework that enhances the model's ability to detect stances on unknown targets by constructing positive and negative samples for contrastive learning.

\textbf{(2) KEprompt \cite{huang2023knowledge}.} This framework designs a knowledge-enhanced prompt tuning method for stance detection. KEprompt includes an automatic vocabulary (AutoV) and background knowledge injection (BKI).

\textbf{(3) TATA \cite{13}.} This model achieves stance detection by combining topic-aware (TAW) and topic-agnostic (TAG) embedding layers.

\textbf{(4) KASD \cite{14}.} This model proposes combining Wikipedia knowledge for retrieval-enhanced generation.

\textbf{(5) MB-Cal \cite{15}.} This model designs a new gated calibration network to mitigate the bias of LLMs in stance detection results.

\textbf{(6) CoSD \cite{16}.} This framework utilizes contrastive heterogeneous topic graph learning to learn the topic-aware semantics and collaborative signals between texts, topics, and stance labels to enhance stance detection.

\textbf{(7) COLA \cite{17}.} This model designs a collaborative role-injection framework involving multiple LLMs, where LLMs are designated different roles, including multi-dimensional text analysis phase, reasoning-enhanced debate phase, and stance conclusion phase. This framework does not require additional labeled data and only needs interaction with pre-trained LLMs, making it highly usable.

\section{Experimental Results}

\begin{table}[!htb]
\centering
\footnotesize
\setlength{\tabcolsep}{2pt} 
\caption{Zero-shot stance detection experiment results}
\begin{tabular}{@{}lcccccc@{}}
\toprule
 & \textbf{HC} & \textbf{FM} & \textbf{LA} & \textbf{A} & \textbf{CC} & \textbf{Avg}  \\
\midrule
\textbf{Baseline models}\\
\addlinespace[4pt]
JointCL & 54.80 & 53.80 & 49.50 & 54.50 & 39.70 & 50.46 \\
\addlinespace[4pt]
TATA & 65.40 & 66.90 & 62.90 & 52.10 & 41.60 & 57.78\\
\addlinespace[4pt]
KASD-LLaMA2 & 77.70 & 65.57 & 57.07 & - & - & - \\
\addlinespace[4pt]
KASD-ChatGPT & 80.32 & 70.41 & 62.71 & 63.95 & 55.83 & 66.64\\
\addlinespace[4pt]
COLA & 81.70 & 63.40 & 71.00 & 70.80 & 65.50 & 70.48\\
\addlinespace[4pt]
LLaMA2-MB-Cal & 75.47 & 73.25 & 67.76 & 64.83 & 58.23 & 67.91 \\
\addlinespace[4pt]
GPT3.5-MB-Cal & 78.50 & \textbf{74.99} & 66.08 & 66.87 & 67.22 & 70.73 \\
\midrule
\textbf{Chain of Stance(ours)}\\
\addlinespace[4pt]
Mistral & \textbf{86.18} & 74.93 & 72.89 & \textbf{77.52} & 70.61 & \textbf{76.43}\\
\addlinespace[4pt]
Qwen1.5 & 78.57 & 72.26 & 70.23 & 73.70 & 63.19 & 71.59\\
\addlinespace[4pt]
LLaMA2 & 77.45 & 70.08 & \textbf{74.10} & 76.11 & 70.85 & 73.72\\
\addlinespace[4pt]
LLaMA3 & 82.90 & 73.51 & 71.39 & 73.84 & \textbf{79.36} & 76.20\\
\bottomrule
\end{tabular}
\label{table2}
\end{table}

\begin{table}[!htb]
\centering
\footnotesize
\setlength{\tabcolsep}{2pt} 
\caption{Few-shot stance detection experiment results}
\begin{tabular}{@{}lcccccc@{}}
\toprule
 & \textbf{HC} & \textbf{FM} & \textbf{LA} & \textbf{A} & \textbf{CC} & \textbf{Avg}  \\
\midrule
\textbf{Baseline models}\\
\addlinespace[4pt]
KEprompt & 77.10 & 68.30 & 70.30 & - & - & - \\
\addlinespace[4pt]
KASD-LLaMA2 & 77.89 & 67.29 & 52.00 & - & - & -\\
\addlinespace[4pt]
KASD-ChatGPT & 80.92 & 70.37 & 63.26 & 61.92 & 62.72 & 67.84 \\
\addlinespace[4pt]
CoSD & 76.35 & 68.96 & 77.29 & \textbf{81.02} & 68.33 & 74.39\\
\addlinespace[4pt]
LLaMA2-MB-Cal & 82.19 & 75.74 & 73.50 & 69.57 & 76.96 & 75.59\\
\addlinespace[4pt]
GPT3.5-MB-Cal & 83.03 & 75.57 & 69.98 & 75.19 & \textbf{84.55} & 77.67 \\
\midrule
\textbf{Chain of Stance(ours)}\\
\addlinespace[4pt]
Mistral & \textbf{87.04} & \textbf{77.33} & \textbf{77.47} & 78.14 & 79.24 & \textbf{79.84}\\
\addlinespace[4pt]
Qwen1.5 & 82.11 & 72.98 & 76.11 & 79.35 & 74.41 & 76.99\\
\addlinespace[4pt]
LLaMA2 & 83.68 & 73.87 & 73.50 & 73.62 & 69.72 & 74.88\\
\addlinespace[4pt]
LLaMA3 & 85.95 & 73.69 & 72.34 & 74.43 & 78.86 & 77.05\\
\bottomrule
\end{tabular}
\label{table3}
\end{table}

\subsection{Main Results}
We conducted both zero-shot and few-shot experiments on the SemEval-2016 dataset. The experimental results are shown in Table\ref{table2} and Table\ref{table3}. For the few-shot experiments, we set the number of shots to 4. 
Based on the experimental results, we can draw the following conclusions: (1):After applying CoS, our models achieved state-of-the-art results in both zero-shot and few-shot settings. Not only did we surpass baseline models, but our results also outperformed the latest models and methods. This demonstrates that CoS can effectively and significantly enhance the stance detection capabilities of LLMs. (2):Except for the LLaMA3-8b model, all other models we used were 7b models. Notably, even though LLaMA2-MB-Cal uses a much larger 70b model, our LLaMA2-7b model still achieved leading results. This highlights the efficiency and effectiveness of CoS, proving that significant improvements can be made without requiring an excessively large model. (3):All four models we tested showed significant improvement after applying CoS. This indicates that CoS is broadly applicable across different LLMs.

\subsection{Error Analysis}
When using the chain of stance (CoS), we conducted an error analysis on the four models. The results are illustrated in Fig.\ref{fig:img3} We categorized the errors into four types: 
\begin{figure}
    \centering
    \includegraphics[width=1\linewidth]{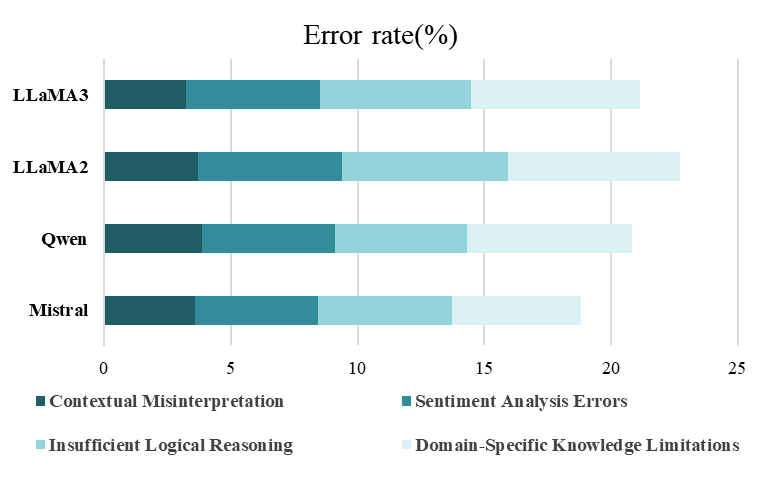}
    \caption{Error Analysis.}
    \label{fig:img3}
\end{figure}

(1)Contextual Misinterpretation: The model may fail to accurately capture key background information or contextual cues within the text, leading to a misinterpretation of the overall meaning. This includes, but is not limited to, misunderstandings of cultural or historical context, misuse of specific terms or slang, and so forth. 
(2)Sentiment Analysis Errors: Even with a correct understanding of the text content, the model might misinterpret the sentiment or tone expressed by the author, thereby affecting stance determination. This is particularly relevant for handling complex emotions like sarcasm or irony. 
(3)Insufficient Logical Reasoning: When the task requires logical reasoning to ensure stance consistency and validity, the model might make incorrect judgments due to a lack of deep understanding or reasoning capabilities. 
(4)Domain-Specific Knowledge Limitations: For specialized domains or specific topics, the model might struggle to accurately determine stances due to insufficient domain knowledge.

By analyzing these error categories, we can identify areas where the chain of stance method can be further improved to enhance the overall accuracy and robustness of stance detection in large language models. 
\subsection{Ablation Study}
We conducted ablation experiments to demonstrate the effectiveness of our method: "w/o CoS" indicates the model without using chain of stance. The results are shown in Table \ref{table4}. The experimental results show that each model achieved significant improvement in detection results after applying our method. 
\begin{table}[!htb]
\centering
\footnotesize
\setlength{\tabcolsep}{3pt} 
\caption{Experimental results of ablation study}
\begin{tabular}{@{}lcccccc@{}}
\toprule
  \textbf{Model} & \textbf{HC} & \textbf{FM} & \textbf{LA} & \textbf{A} & \textbf{CC} & \textbf{Avg}  \\
\midrule
\addlinespace[4pt]
\rowcolor{gray!25} Mistral-CoS & 86.18 & 74.93 & 72.89 & 77.52 & 70.61 & \textbf{76.43} \\
\addlinespace[4pt]
w/o CoS & 79.80 & 70.41 & 71.08 & 74.39 & 57.63 & 70.66 \\
\addlinespace[4pt]
\rowcolor{gray!25} Qwen-CoS & 78.57 & 72.26 & 70.23 & 73.70 & 63.19 & \textbf{71.59}\\
\addlinespace[4pt]
w/o CoS & 74.87 & 68.70 & 56.58 & 67.02 & 50.55 & 63.54 \\
\addlinespace[4pt]
\rowcolor{gray!25} LLaMA2-CoS & 77.45 & 70.08 & 74.10 & 76.11 & 70.85 & \textbf{73.72}\\
\addlinespace[4pt]
w/o CoS & 70.94 & 63.69 & 59.36 & 52.56 & 43.65 & 58.04\\
\addlinespace[4pt]
\rowcolor{gray!25} LLaMA3-CoS & 82.90 & 73.51 & 71.39 & 73.84 & 79.36 & \textbf{76.20} \\
\addlinespace[4pt]
w/o CoS & 78.52 & 70.00 & 67.86 & 67.87 & 65.49 & 69.95\\
\bottomrule
\end{tabular}
\label{table4}
\end{table}

\begin{figure}
    \centering
    \includegraphics[width=1\linewidth]{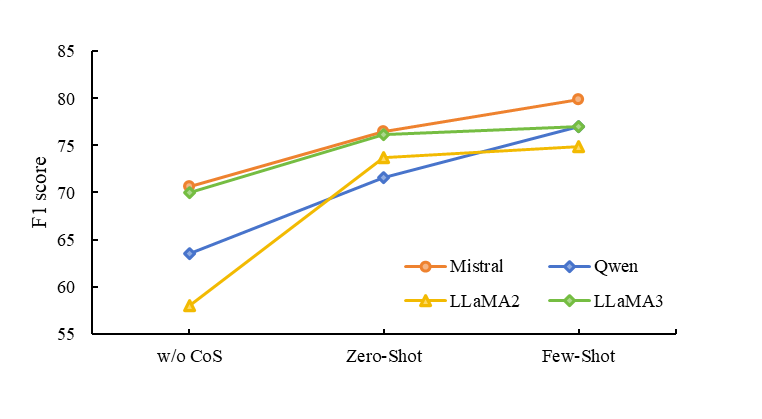}
    \caption{Performance Variations Across Different Models.}
    \label{fig:img4}
\end{figure}

Additionally, we compared the F1 score changes of the four models (Mistral, Qwen, LLaMA2, and LLaMA3) under three different scenarios: without CoS, zero-shot learning, and few-shot learning. The comparison results are shown in Fig.\ref{fig:img4} The results clearly demonstrate that the accuracy of stance detection improved significantly across all models when using CoS. Specifically, the Mistral model performed the best in all scenarios, with its F1 score increasing from approximately 70 without CoS to about 76.43 in zero-shot learning, and further to about 79.84 in few-shot learning. The other models (Qwen, LLaMA2, and LLaMA3) showed similar trends, although their absolute scores were slightly lower than those of the Mistral model. These results indicate that the adoption of the CoS strategy significantly enhances the performance of models in stance detection tasks. 

\section{Conclusion and Future Work}

To explore the potential of Large Language Models (LLMs) in human stance detection, we propose a novel method called chain of stance (CoS) in this paper. This approach explicitly models the reasoning process of stance detection. By capturing a sequence of stance-related assertions and their supporting evidence, CoS offers more interpretable and transparent predictions compared to traditional classification methods. Our experimental results demonstrate that CoS significantly outperforms all baselines, greatly enhancing the accuracy of LLMs in stance detection tasks. In future work, we will continue to explore how CoS can fully leverage the potential of LLMs. 

\section*{Limitations}
The important limitation of this work is that the length of the prompts leads to a decrease in the model's computational efficiency. Although computational efficiency is not within the scope of this paper, improving computational efficiency while enhancing the model's detection accuracy will be our goal in future work.

\bibliography{references}
\bibliographystyle{acl_natbib}

\end{document}